\documentclass[10pt,twocolumn,letterpaper]{article}

\usepackage{cvpr}
\usepackage{times}
\usepackage{epsfig}
\usepackage{graphicx}
\usepackage{amsmath}
\usepackage{amssymb}

\usepackage{float}

\newcommand{\R}{\mathbb{R}}
\newcommand{\E}{\mathbb{E}}

\newcommand{\tr}[1]{#1^{\mathrm{T}}}
\newcommand{\paren}[1]{\left(#1\right)}

\usepackage[pagebackref=true,breaklinks=true,letterpaper=true,colorlinks,bookmarks=false]{hyperref}

 \cvprfinalcopy 

\def\cvprPaperID{6037} 

\ifcvprfinal\fi
\begin{document}

\title{Understanding the (un)interpretability of natural image distributions using generative models}

\author{Ryen Krusinga \\
{\tt\small krusinga@cs.umd.edu}
\and
Sohil Shah \\
{\tt\small sohilas@terpmail.umd.edu}
\and
Matthias Zwicker \\
{\tt\small zwicker@cs.umd.edu}
\and
Tom Goldstein \\
{\tt\small   \hspace{2mm} tomg@cs.umd.edu  \hspace{2mm}}
\and
David Jacobs \\
{\tt\small djacobs@umiacs.umd.edu}
}

\maketitle

\begin{abstract}


Probability density estimation is a classical and well studied problem, but standard density estimation methods have historically lacked the power to model complex and high-dimensional image distributions.  More recent generative models leverage the power of neural networks to implicitly learn and represent probability models over complex images.  We describe methods to extract explicit probability density estimates from GANs, and explore the properties of these image density functions.  We perform sanity check experiments to provide evidence that these probabilities are reasonable.  However, we also show that density functions of natural images are difficult to interpret and thus limited in use.  We study reasons for this lack of interpretability, and show that we can get interpretability back by doing density estimation on latent representations of images.  

\end{abstract}

\begin{figure}[!htb]
    \centering

    \setlength{\tabcolsep}{2pt} 
    \begin{tabular}{c c}
        \includegraphics[width=0.5\columnwidth]{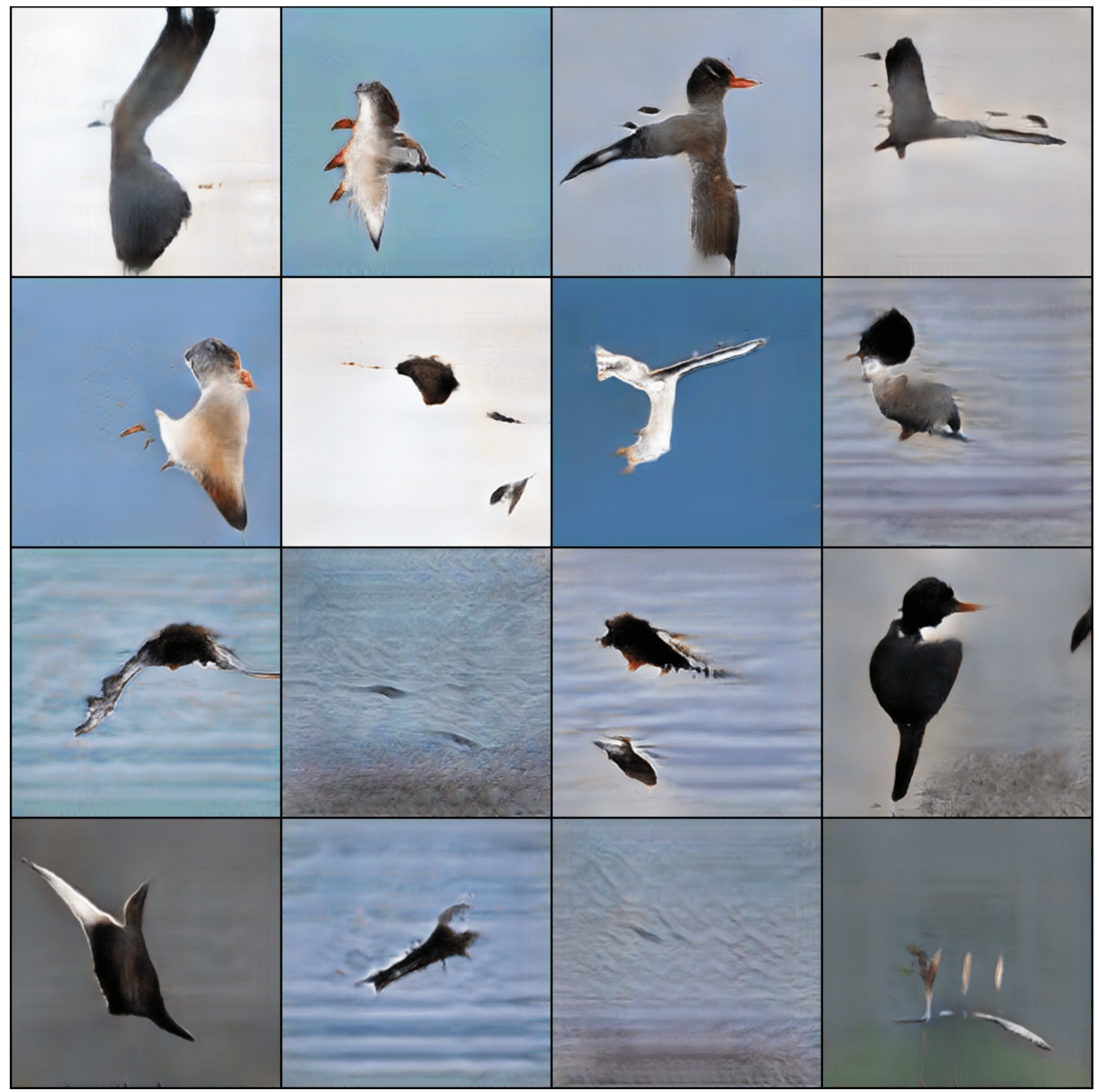} & 
        \includegraphics[width=0.5\columnwidth]{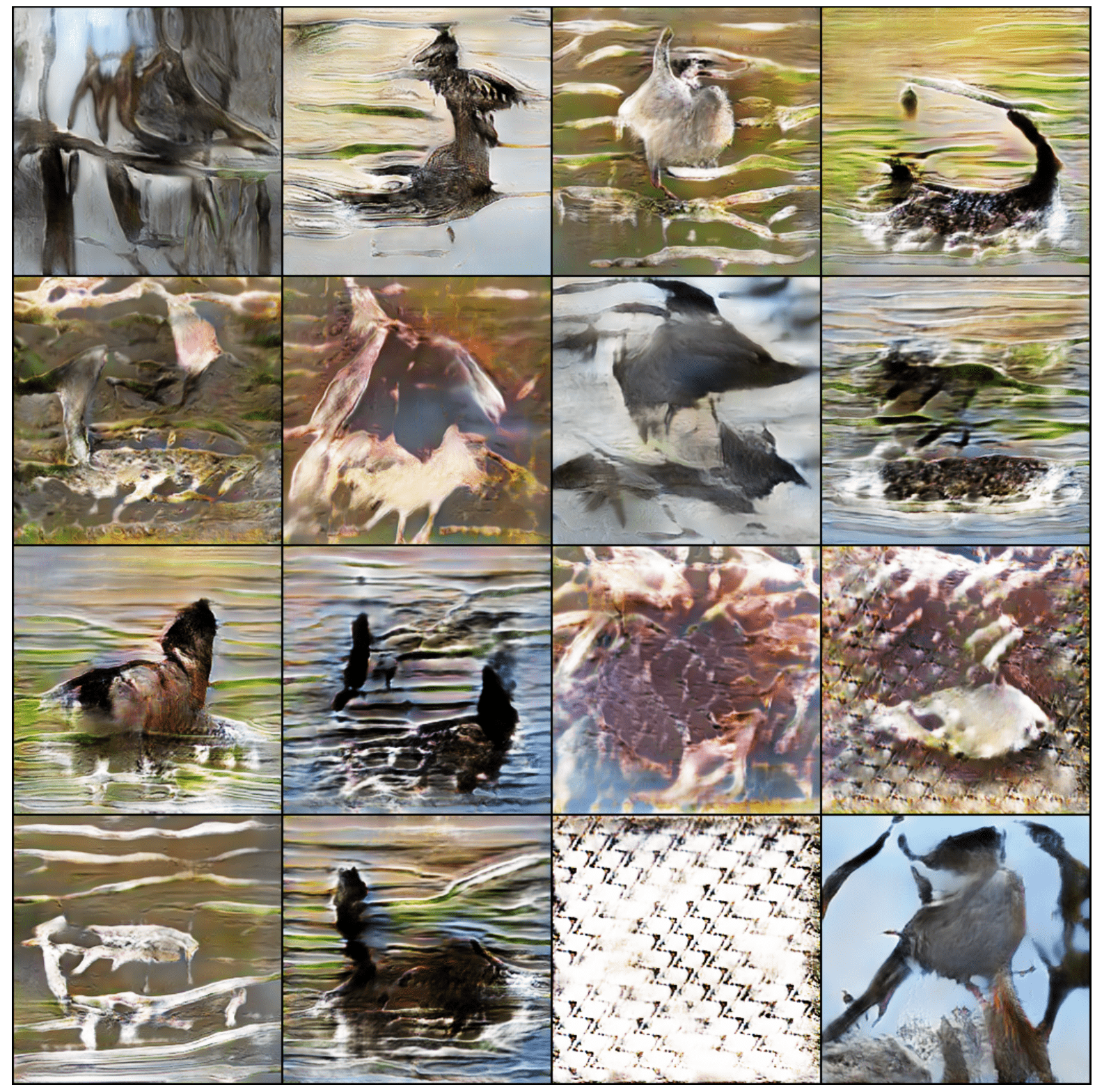} \\
        \includegraphics[width=0.5\columnwidth]{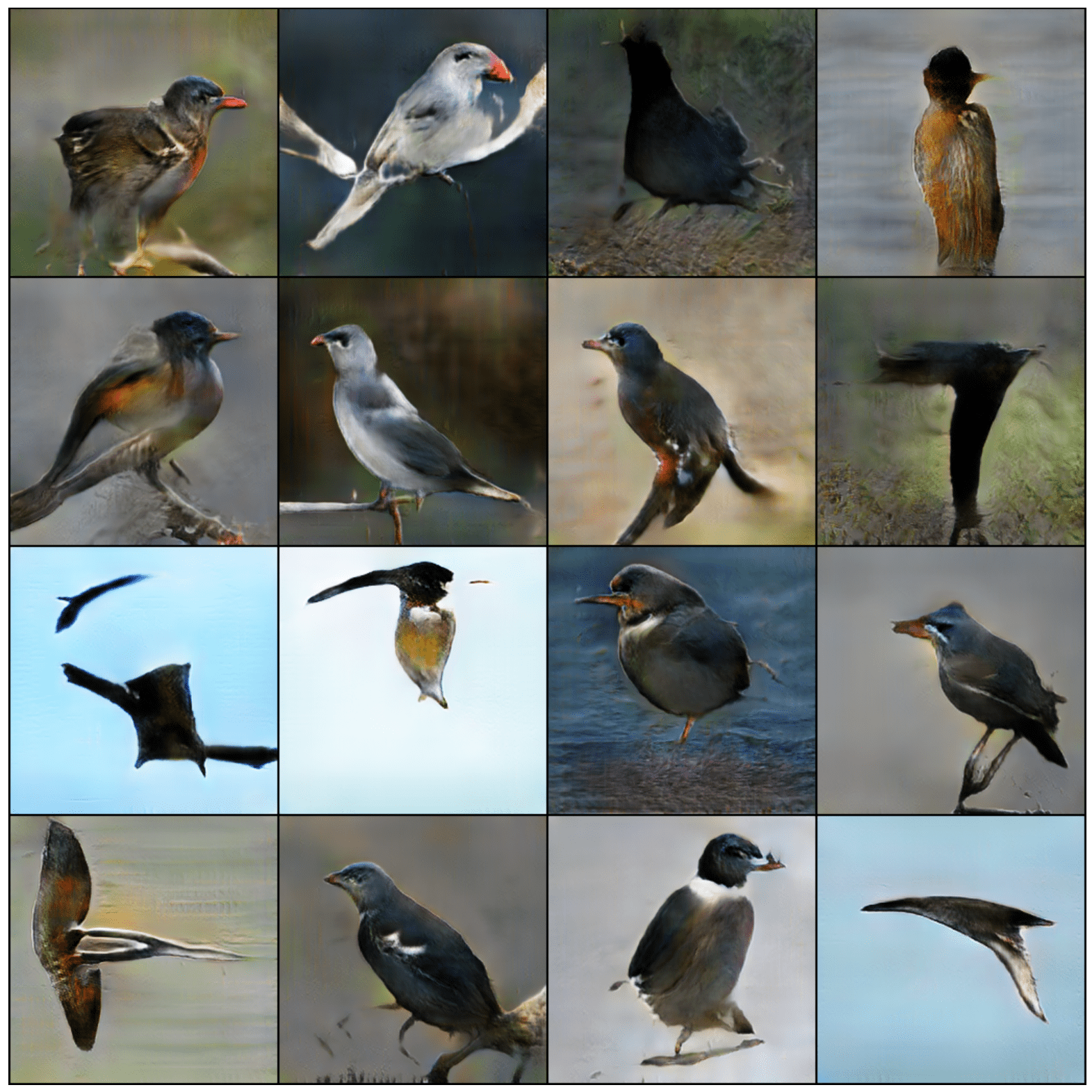} &
        \includegraphics[width=0.5\columnwidth]{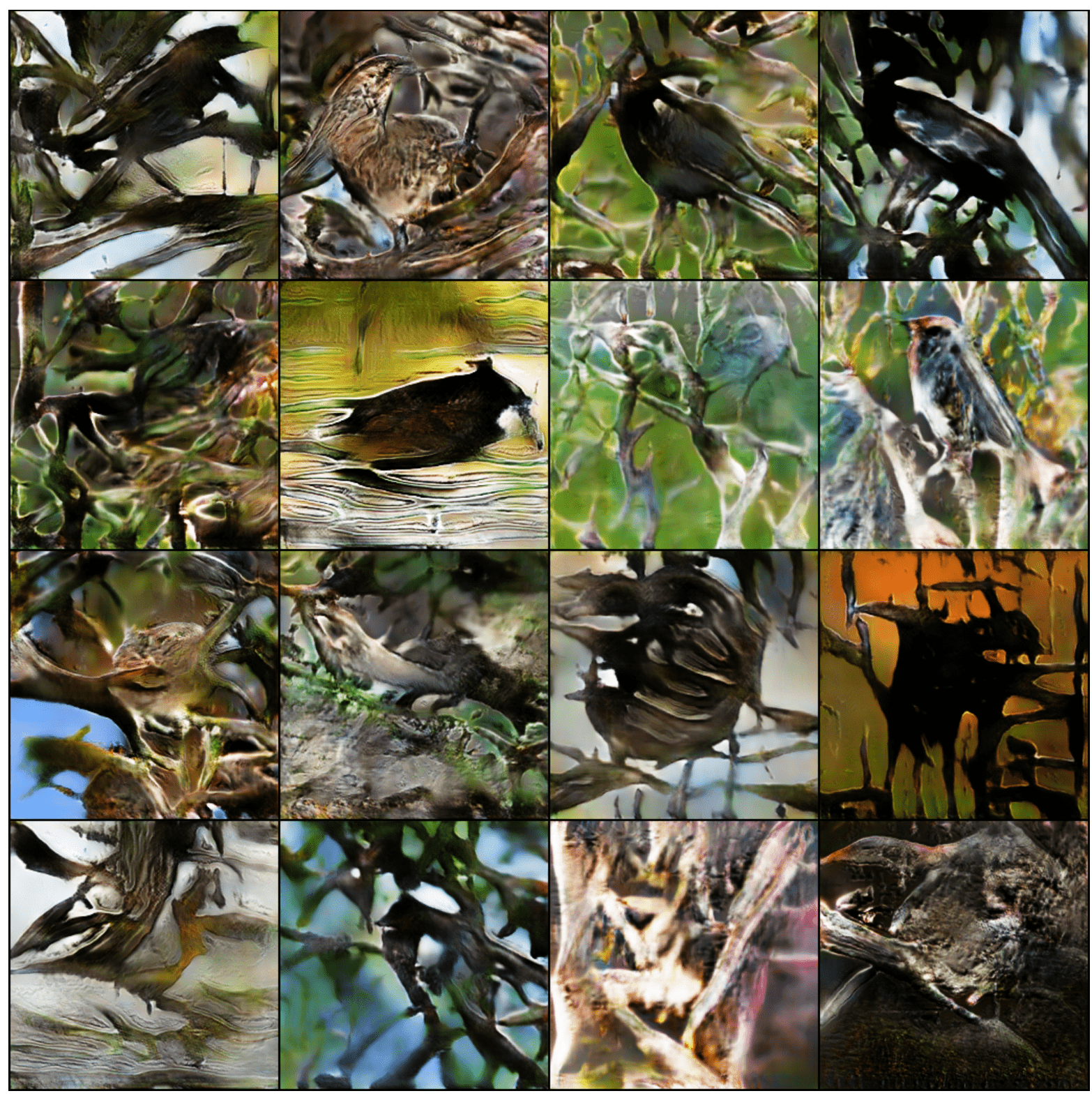}  \\
    \end{tabular}
    \caption{StackGAN images with high probability density are typical in appearance, while low density images appear as distribution ``outliers'' with strange behaviors.  Top left: Most likely StackGAN samples for Caption 1: ``A bird with a very long wing span and a long pointed beak.''  Top right: least likely StackGAN samples for Caption 1. Bottom left: most likely stackGAN samples for Caption 2: ``This bird has a white eye with a red round shaped beak.''  Bottom right: least likely stackGAN samples for Caption 2.   }
	\label{fig:stackgan}
\end{figure}

\section{Introduction}


Researchers have long sought to estimate probability density functions (PDFs) of images.  These generative models can be used in problems such as image synthesis, outlier detection, as priors for image restoration, and in approaches to classification.  There have been some impressive successes, including building generative models of textures for texture synthesis, and using low-level statistical models for image denoising.  However, building accurate densities of full, complex images remains challenging.

Recently there has been a flurry of activity in building deep generative models of complex images, including the use of GANs to generate stunningly realistic complex images.  While some deep models focus explicitly on building probability densities of images, the emphasis has been on GANs that generate the most realistic imagery.  Implicitly, though, these GANs also encode probability densities.  In this paper we explore the question of whether these implicit densities effectively capture the intuition of a probable image.  We show that in some sense the answer is ``no".  But, we show that by computing PDFs over latent representations of images, we can do better. 

We first propose some methods for extracting probabilities from GANs.  It is well known that when a bijective function maps one density to another, the relationship between the two densities can be understood using the determinant of the Jacobian of the function.  GANs are not bijective, and map a low-dimensional latent space to a high-dimensional image space.  In this case, we modify the standard formula so that we can extract the probability of an image given its latent representation.  This allows us to compute probabilities of images generated by the GAN,  which we then use to train a regressor that computes probabilities of arbitrary images.

We perform sanity checks to ensure that GANs do indeed produce reasonable probabilities on images.  We show that GANs produce similar probabilities for training images and for held out test images from the same distribution.  We also show that when we compute the probability of either real or generated images, the most likely images are of high quality, and the least likely images are of low quality. An example of this last result is shown in Figure \ref{fig:stackgan}, which displays the most and least likely caption-conditioned images generated by a StackGAN \cite{zhang2017stackgan} trained on the CUB dataset~\cite{WelinderEtal2010}.

Researchers have sought for decades to interrogate complex and high dimensional image distributions, and with the power of deep generative models we finally can.  Unfortunately, now that we have these tools at our disposal, we show that probability densities learned on images are difficult to interpret and have unintuitive behaviors.  The most likely images tend to contain small objects with large, simple backgrounds, while images with complex backgrounds are deemed unlikely.   For example, if we train a GAN on MNIST digits and then test on new MNIST digits, all the most likely digits are simple 1s.  If we take all the 1s out of the training set, then when we test on the full set of MNIST digits including 1s, the 1s are {\em still} the most likely, even though the GAN never saw them during training.  In fact, even if we train a GAN on CIFAR images of real objects, the GAN will produce higher probabilities for MNIST images of 1s than for most of the CIFAR images.   We investigate these strange properties of density functions in detail, and explore reasons for this lack of interpretability. 

Fortunately, we can mitigate this problem by doing probability density estimation on the latent representations of the images, rather than their pixel representations.  This approach gives us back interpretability; we obtain probability distributions with inliers and outliers that coincide with our intuition.

In parallel to our work, \cite{nalisnick2018deep} also addresses the interpretability of density functions over images, claiming that seemingly uninterpretable density estimates result from inaccurate estimation on out-of-sample images \cite{nalisnick2018deep}.  Our thesis is different, as we argue that density estimation is often accurate even for unusual images, but the true underlying density function (even if known exactly) is fundamentally difficult to interpret.

\section{Background}

There are many classical models for density estimation in low-dimensional spaces.  Non-parametric methods such as Kernel density estimation (i.e., Parzan windows \cite{parzen1962estimation,heidenreich2013bandwidth}) can model simple distributions with light tails, and nearest-neighbor classifiers (eg., \cite{boiman2008defense} ) implicitly use this representation.  Directed graphical models (eg., Chow-Liu trees and related models \cite{chow1968approximating}) have also been used for classification \cite{mattarimproved}.  However, these models do not scale up to the complexity or dimensionality of image distributions. 

There is a long history of approximating the PDFs of images using simple statistical models.   These approaches succeed at estimating some low-dimensional marginal distribution of the true image density.  Modeling the complete, high-dimensional distribution of complex images is a substantially more difficult problem.  
 For example, \cite{olshausen1996natural} models the low-level statistics of natural images.  \cite{portilla2003image} uses conditional models on the wavelet coefficients of images and shows that these models can improve image denoising.  \cite{roth2005fields} learns and applies image priors based on Fields of Experts.  Markov models have also been used to synthesize textures with impressive realism \cite{de1998texture,efros1999texture}. 


Neural networks have been used to build generative models of images.  \cite{park2016image,timofte2012naive} do so assuming independence of pixels or patches.  Restricted Boltzmann Machines \cite{smolensky1986information} and Deep Boltzmann machines \cite{salakhutdinov2010efficient} also model image densities.  However these methods suffer from complex training and sampling procedures due to mean field inference and expensive Monte Carlo Markov Chain methods \cite{salimans2015markov}. Another approach, Variational Autoencoders \cite{kingma2013auto}, simultaneously learn a generative model and an approximate inference, and offer a powerful approach to modeling image densities.  However, they tend to produce blurry samples and are limited in application to low-dimensional deep representations.

Recently, GANs\cite{goodfellow2014generative}  have presented a powerful new way of building generative models of images with remarkably realistic results \cite{brock2018large}.  
Generative adversarial networks are neural network models trained adversarially to learn a data distribution. They consist of a generator $G_{\theta}:\R^n \rightarrow \R^m$ and a discriminator $D_{\phi}: \R^m \rightarrow \R$, where $n$ is the dimension of a latent space with probability distribution $P_z$ and $m$ is the dimension of the data distribution $P_d$, which is equal to width $\times$ height $\times$ \#colors in the case of images. In the original GAN, the discriminator produces a probability estimate as output, and the GAN is trained to reach a saddle point via the learning objective
\begin{multline}
\min_{\theta} \max_{\phi} \E_{x \sim P_d}[\log D_{\phi}(x)]\\ + 
\E_{z\sim P_z}[\log (1-D_{\phi}(G_{\theta}(z))],
\end{multline}
which incentivizes the generator to produce samples that the discriminator classifies as likely to be real, and the discriminator to assign high probability values to real points and low values to fake points. Unfortunately, GANs don't produce explicit density models -- the GAN is capable of sampling the density, but not evaluating the density function directly.  

A major limitation of GANs is that they are not invertible.  So, given an image, one does not have access to its latent representation, which could be used to calculate the image's probability.
To overcome this problem, Real Non-Volume-Preserving transformations (Real NVP) \cite{dinh2016density} learn an invertible transformation from the latent space to images.  This yields an explicit probability distribution in which exact likelihood values can be computed. Real NVP can be trained using either maximum likelihood methods or adversarial methods, or a combination of both, as in FlowGAN \cite{grover2017flow}. Both of these models have proven effective at generating high-quality images. (Related models: \cite{dinh2014nice}, \cite{papamakarios2017masked}).

In this paper, we choose to focus on the use of non-invertible GANs to estimate image probability.  One issue with invertible GANs is that the latent space must be of the same dimension as the image space, which becomes problematic for large, high-dimensional images.  Also, non-invertible GANs currently produce the highest quality images, suggesting that they implicitly represent the most accurate probability distributions.  Furthermore, non-invertible GANs use simpler network architectures and training procedures. The standard DCGAN \cite{radford2015unsupervised}, for example, consists of basic convolutional layers with batch norm and ReLU transformations. By contrast, Real NVP requires a scheme of coupling, masking, reshaping, and factoring over variables.  Our proposed methods can be applied to any GAN, so that they can leverage any improvements made as new GAN architectures come along.  

Extracting density estimates from GANs presents several challenges.  A  (non-invertible) GAN learns an embedding of a lower-dimensional latent space (the random codes) into a much higher dimensional space (the space of all possible images of a certain size). Thus, the probability distribution that it learns is restricted to a low-dimensional manifold within the higher-dimensional space. Exact probabilities for images can be computed via the Jacobian if the latent code is known, as we will show in the next section, but probabilities are technically zero for images that are not exactly generated by any latent code. Extending probabilities meaningfully beyond the data manifold requires either incorporating an explicit noise model, such as in the recent Entropic GAN \cite{balaji2018entropic}, or learning a projection from images to latent codes, such as in BiGAN \cite{donahue2016adversarial}. 

In this paper, we avoid these complexities by creating a simple regressor network that accepts an image and returns its estimated probability density.  Training such a regressor network is easy if one has a large dataset of images labeled with their probability densities.  In section \ref{extract} we describe a simple method for obtaining such a dataset.



\section{Extracting probability densities} \label{extract}
A GAN generator $G$ produces an image $G(Z)$ from the random variable $Z$ with a know latent distribution $P_z$. But what is the distribution of the induced random variable $G(Z)$? If $G$ is differentiable and bijective, then the change of variables formula \cite{munkres2018analysis} provides a method for determining the exact density of the warped distribution. For $x = G(z)$ we have
\begin{equation}
P_d(x) = P_z(G^{-1}(x))| \det \partial G^{-1}(x) |
\end{equation}
where $\partial G^{-1}$ is the Jacobian of the inverse function and $|\cdot|$ is the determinant. Intuitively, the determinant of the Jacobian measures the change in volume of a unit cube mapped from the target to the latent distribution, and P(z) measures the probability of the corresponding volume in the latent distribution. This corresponds to our intuition that the probability of an event in the target space is equal to the probability of everything in the latent space that maps to the event.


But there is an immediate problem. Most GAN generators are not bijective.  What's worse, they map a low-dimensional latent space to a high-dimensional pixel space, so the Jacobian is not square and we cannot compute a determinant. The solution is to perform calculations not on the codomain, but on the low-dimensional manifold consisting of the image of the latent space under G. If G is differentiable and injective, then this manifold has the same intrinsic dimensionality 
as the latent space, and we can consider how a unit cube in the $n$-dimensional latent space distorts as it maps onto the (also $n$-dimensional) image manifold. The resulting modified formula is
\begin{equation}
\label{eq:jacob}
P(x) = P(z)|\det \tr{\partial G}(z)\partial G(z)|^{-\frac{1}{2}}.
\end{equation}
The formula uses the fact that $\det(M^{-1}) = (\det M)^{-1}$ for any square matrix $M$.  It also uses the fact that the squared volume of a parallelpiped in a linear subspace is computed by projecting to subspace coordinates via the transpose of the coordinate matrix, resulting in the square matrix $\tr{\partial G}\partial G$ (an expression which is known as a \emph{metric tensor}), and then taking the determinant.

The Jacobian $\partial G$ can be computed analytically from the network computation graph, or numerically via a finite difference approximation. Once computed, we can find the above determinant via a QR decomposition. If $\partial G = Q\cdot R$, where $Q$ is an $m\times n$ matrix with orthonormal columns and R is an $n \times n$ upper-triangular matrix, then
\[
\det \tr{\partial G}\partial G = \det \tr{R}\tr{Q}QR = \det(R)^2 = \paren{\prod_{i=1}^n r_{ii}}^2
\]
Substituting back into equation \ref{eq:jacob}, we obtain the probability formula
\begin{equation}
\label{eq:practical}
P(x) = P(z)\prod_{i=1}^n |r_{ii}|^{-1}.
\end{equation}
In practice, we evaluate the log-density using the formula $\log P(x) = \log P(z)-\sum_{i=1}^n \log(|r_{ii}|)$ rather than the density itself to avoid numerical over/underflow.

To generalize probability predictions to novel images, we train a separate regressor network on samples from $G$, which are labeled with their log-probability densities. This regressor predicts probabilities directly from images. We will refer to this as the \emph{pixel regressor}.

\section{Methods}
  Here we describe the details of the experimental methods; the reader may skip this section without loss of continuity. Subsequent sections of the paper will describe experiments using these methods, including a variant of the regressor that learns PDFs on the latent representations rather than the pixel representations of the images.

The entire density estimation process follows this simple template:
\begin{enumerate}
\item Train a GAN on a dataset (MNIST, CIFAR).
\item Sample the GAN to produce many triplets $(z, G(z), P(G(z))),$ each comprising a latent code, sampled image, and image log-density.
\item Train a regressor network on the samples.  This network maps images to their probability density.
\end{enumerate}
All GAN networks,
were trained with the Adam optimizer using a learning rate of 0.0002 and beta parameters $(0.5, 0.999)$. All regressor networks were trained using Adam with the same beta parameters but a learning rate of 0.0001.

\textbf{DCGAN.}
As the basis of our networks, we use a standard DCGAN architecture \cite{radford2015unsupervised}, consisting of four $4 \times 4$ (de)convolutional layers separated by batch norm and ReLU nonlinearities. Standard ReLU was used for the generator, while the discriminator used LeakyReLU with a negative slope of 0.2. All of our DCGAN models used a 100-dimensional latent code vector sampled from a uniform Gaussian prior distribution.

\textbf{InfoGAN objective.}
When we model the PDFs using the latent variables, we use an InfoGAN-inspired \cite{chen2016infogan} training procedure. The intent of this procedure was to force the latent space to learn a more structured representation of the data, similar to how InfoGAN was able to learn structured representations of MNIST digits (albeit using a different latent code distribution). It is also possible to use the $Q$ and $D$ networks directly as the regressor for latent codes; however, we found that in practice, we could obtain superior performance by training a separate regressor on numerous samples from the fully trained GAN. We will refer to this as the \emph{latent regressor.}

 To implement this objective, we added to DCGAN a simple $Q$-network that takes the output of the third convolutional layer of the discriminator as its input. The Q network consisted of a convolutional layer followed by two linear layers, separated by LeakyReLU nonlinearities. The final linear layer produced a 100-dimensional vector intended to reconstruct the latent z-code that produced the current GAN sample. As per \cite{chen2016infogan}, the loss function on Q should be designed to maximize the mutual information $\mathrm{MI}(Q(z),z)$ of the output of the Q-network (a parameterized probability distribution) with the latent probability distribution. In this case, if we treat the output of Q as the mean of a Gaussian distribution with a fixed identity covariance matrix, then maximizing the mutual information is equivalent to minimizing the squared error between the Q output and the true latent code $z$. Thus the loss function is
\begin{multline}
\min_{\theta, \tau} \max_{\phi} \E_{x \sim P_d}[\log D_{\phi}(x)]\\ + 
\E_{z\sim P_z}[\log (1-D_{\phi}(G_{\theta}(z)) + \lambda ||Q_{\tau}(z)-z||^2]
\end{multline}
where $Q_{\tau}(z)$ represents the output of Q after running $G(z)$ through the discriminator and taking the third convolutional layer as input to Q. The Q network is updated in the same step as G network.


\textbf{Regressor.}
The DCGAN discriminator architecture was also used as the regressor architecture. No sigmoid is used on the output, as we train the regressor to predict log probabilities. We used an L2 loss. We also experimented with a smooth L1 loss and found that this did not affect the qualitative results significantly.




\textbf{Data.}
We scaled the MNIST dataset up to size $32 \times 32 \times 3$ so we could use the same model on MNIST and CIFAR.  This facilitates several experiments below that use both datasets.


\section{Sanity check: do GANs yield accurate probability estimates?}

\begin{figure}[!htb]
    \centering

    \setlength{\tabcolsep}{2pt} 
    \begin{tabular}{c c}
        \includegraphics[width=0.5\columnwidth]{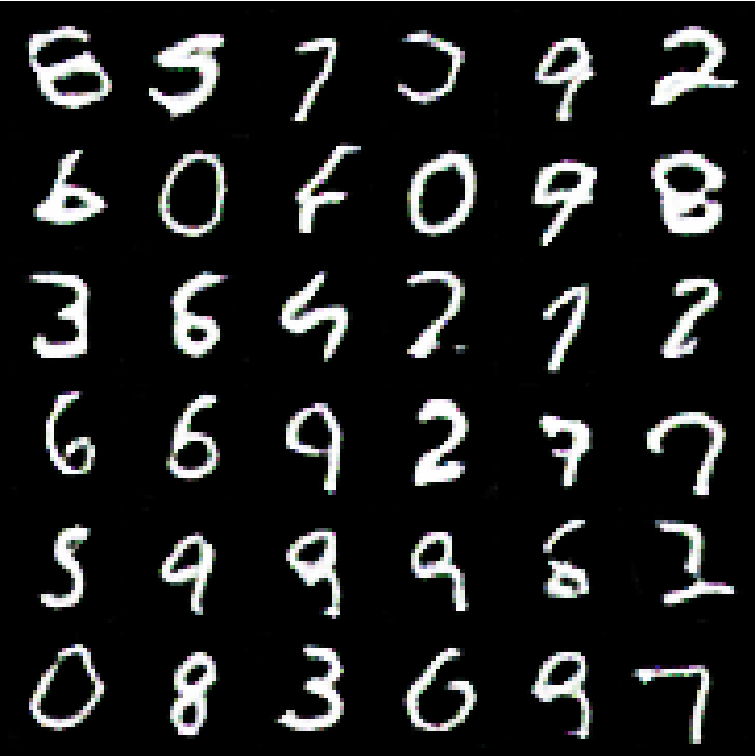} & 
        \includegraphics[width=0.5\columnwidth]{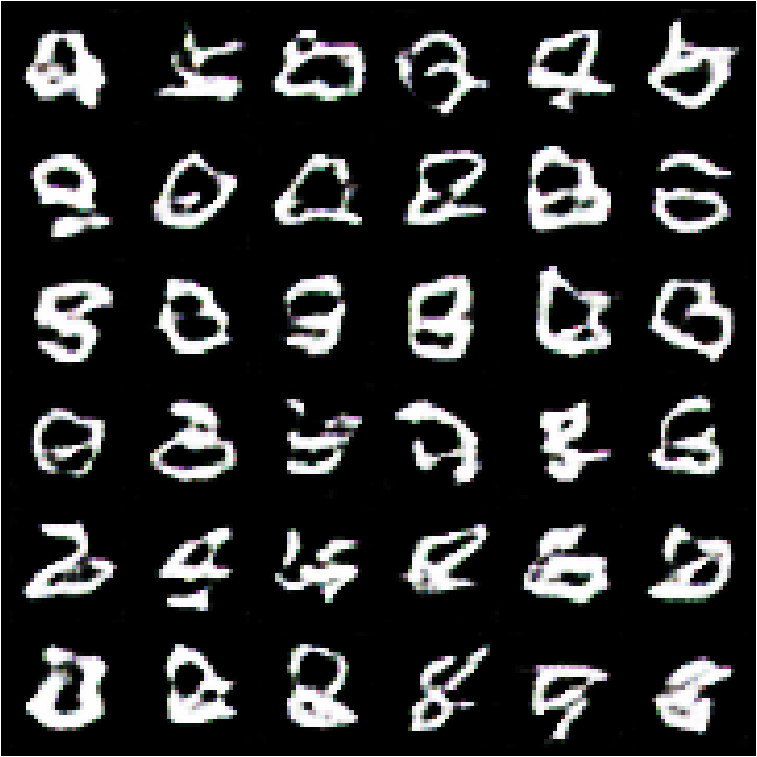} \\
        Random generated &
        Least likely generated 
    \end{tabular}
    \caption{Left: samples from a GAN trained on MNIST. Right: samples of lowest probability according to the pixel regressor.}
    \label{fig:mnist_rand_bottom}
\end{figure}

\begin{figure}[!htb]
    \centering

    \setlength{\tabcolsep}{2pt} 
    \begin{tabular}{c c}
        \includegraphics[width=0.5\columnwidth]{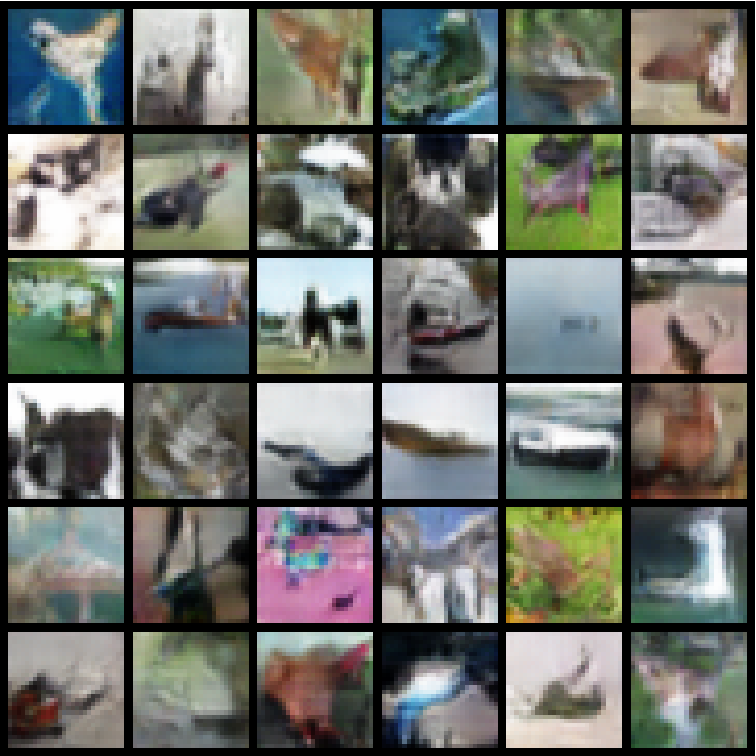} & 
        \includegraphics[width=0.5\columnwidth]{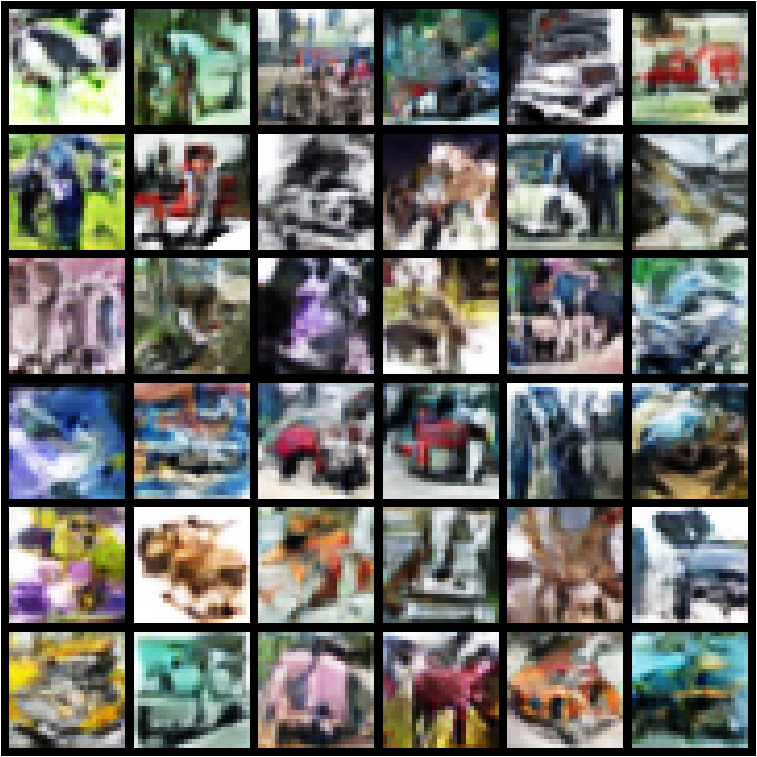} \\
        Random generated &
        Least likely generated  \\
    \end{tabular}
    \caption{Left: samples from a GAN trained on CIFAR. Right: samples of lowest probability according to the pixel regressor.}
    \label{fig:cifar_rand_pixel}
\end{figure}

\begin{figure}
\includegraphics[width=\columnwidth]{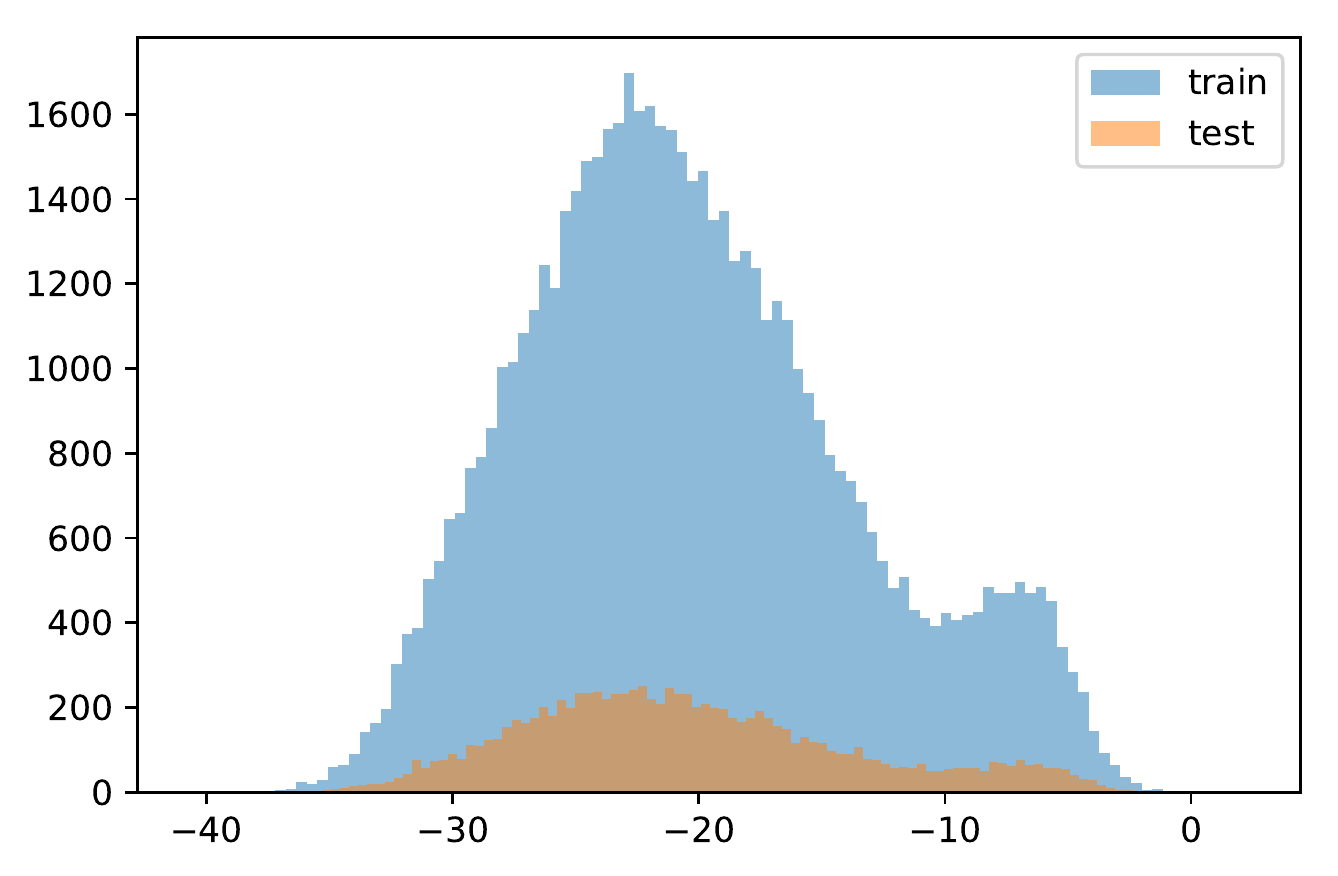}
\caption{Histogram of log probabilities of MNIST train and test data as predicted by a pixel regressor for an MNIST GAN.}
\label{fig:mnist_train_test}
\end{figure}

\begin{figure}
\includegraphics[width=\columnwidth]{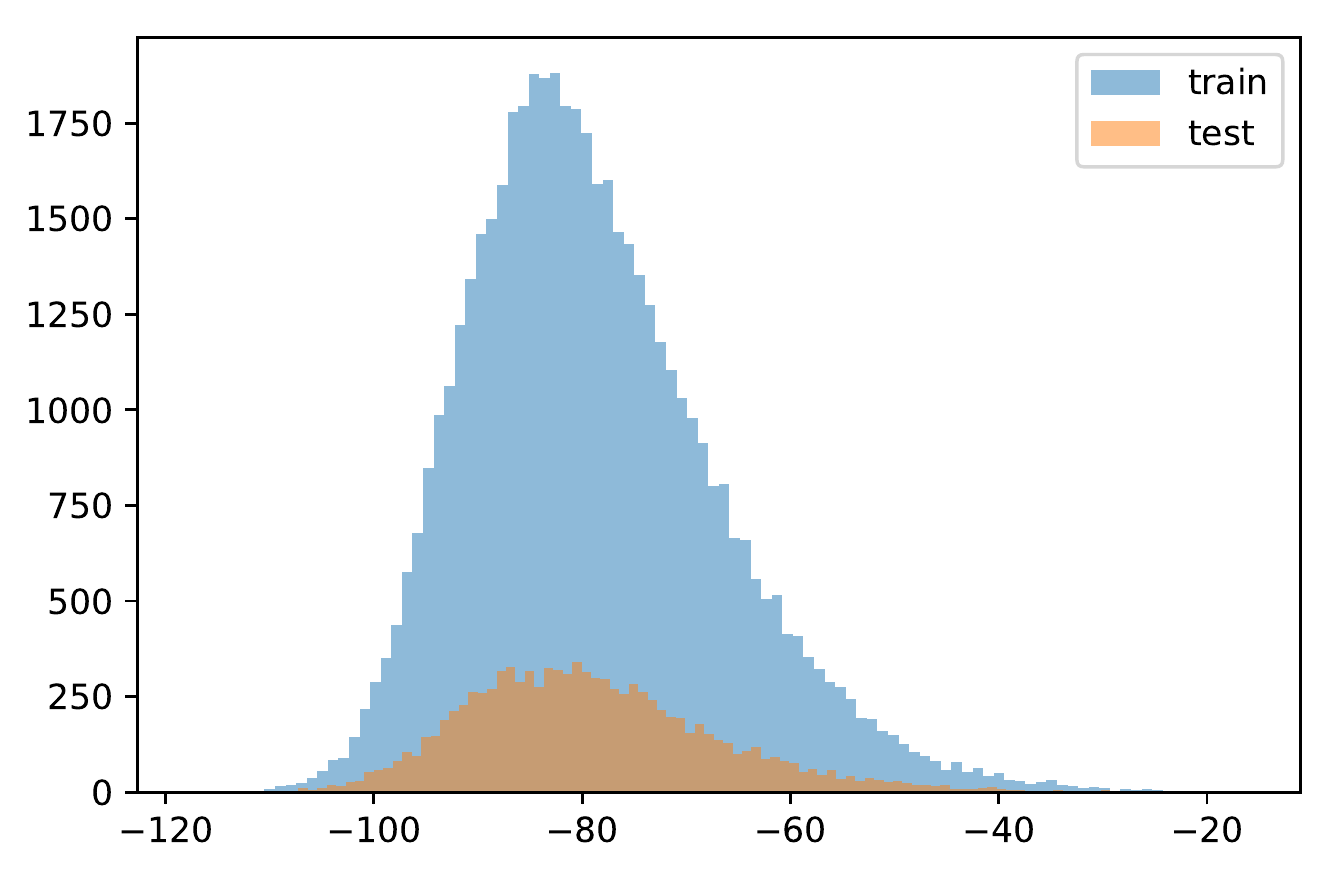}
\caption{Histogram of log probabilities of CIFAR train and test data as predicted by a pixel regressor for a CIFAR GAN.}
\label{fig:cifar_train_test}
\end{figure}

The accuracy of GAN-based density estimation depends on the accuracy of the generated probability labels, and the ability of the regression network to generalize to unseen data.  In this section, we investigate whether the obtained probability densities are meaningful.  We do this quantitatively by comparing histograms of predicted densities in the train and test datasets, and also qualitatively by examining how probability density correlates with image quality. 

\subsection{Comparing histograms} \label{compare}

The GAN and regressor model can be inaccurate because of under-fitting (e.g., missing modes), or overfitting (assigning excessively high density to individual images).  We test for these problems by plotting histograms for the probability densities on both the train and test data to validate that these distributions have high levels of similarity.  

Results are shown in Figures \ref{fig:mnist_train_test} and \ref{fig:cifar_train_test}.  The test histograms appear as a scaled-down version of the train histograms because the test sets contain fewer samples (we did not normalize the histograms by number of samples because this difference in scale helps in seeing both distributions on the same figure).  For both MNIST and CIFAR, we see a very high degree of similarity between test and train distributions, indicating a good model fit (without over-fitting).

\subsection{Visualizing typical and low density images}\label{highlow}
We get a stronger sense for what the density estimator is doing by visualizing ``outliers'' that have low probability density.   Figures \ref{fig:mnist_rand_bottom} and \ref{fig:cifar_rand_pixel} show typical samples produced by the GAN models for MNIST and CIFAR.  We see that the GANs fit the distributions nicely, as typical samples reflect what we want these images to look like.  However, the lowest probability outliers (selected from 50,000 GAN random samples) are extremely irregular and clearly lie away from the modes of the distribution. 

These visualizations suggest that GAN-based density estimators make reasonable density predictions.  However, we will see in the next section that even highly accurate density estimation can have unreasonable consequences for some tasks.

\section{Be careful what you wish for: the difficulties of interpreting image densities} \label{wish}

\begin{figure}[!htb]
    \centering

    \setlength{\tabcolsep}{2pt} 
    \begin{tabular}{c c}
        \includegraphics[width=0.5\columnwidth]{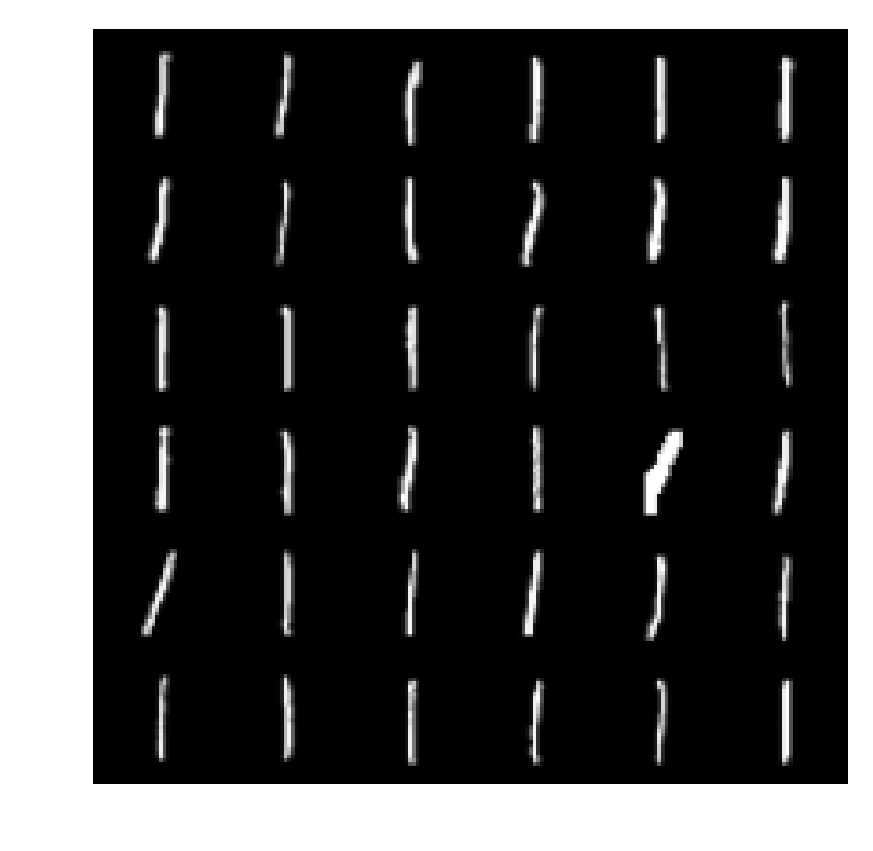} & 
        \includegraphics[width=0.5\columnwidth]{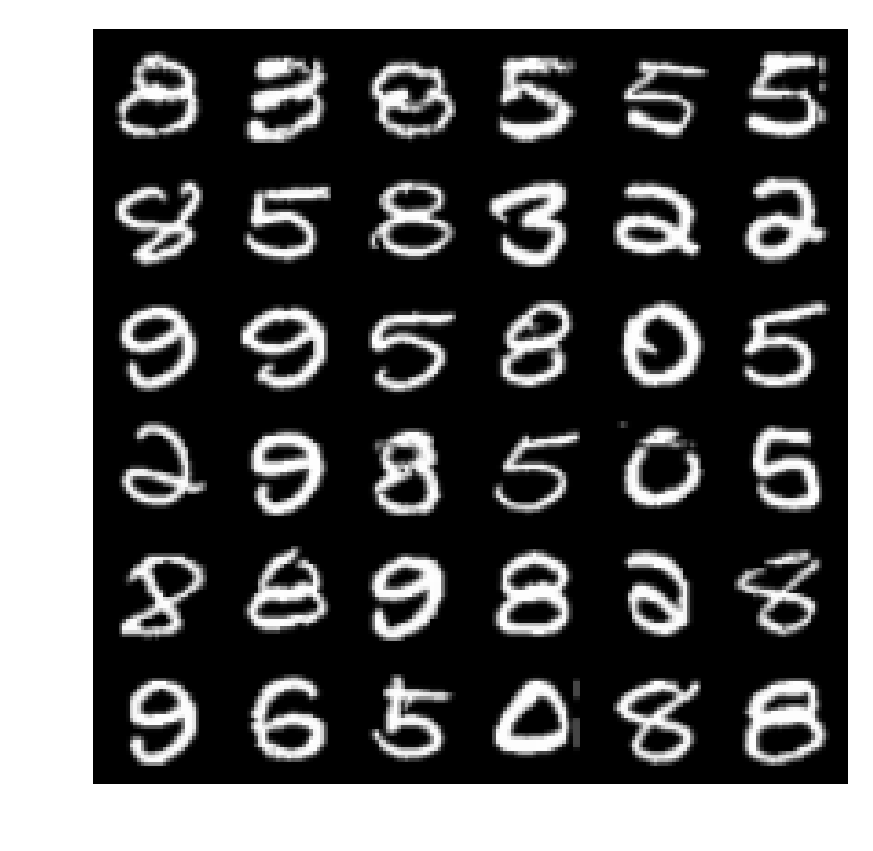} \\
        Most likely MNIST &
        Least likely MNIST
    \end{tabular}
    \caption{Most and least likely MNIST digits as predicted by a probability regressor for a GAN trained on MNIST.}
    \label{fig:mnist_real_top_bottom}
\end{figure}

\begin{figure}[!htb]
    \centering

    \setlength{\tabcolsep}{2pt} 
    \begin{tabular}{c c}
        \includegraphics[width=0.5\columnwidth]{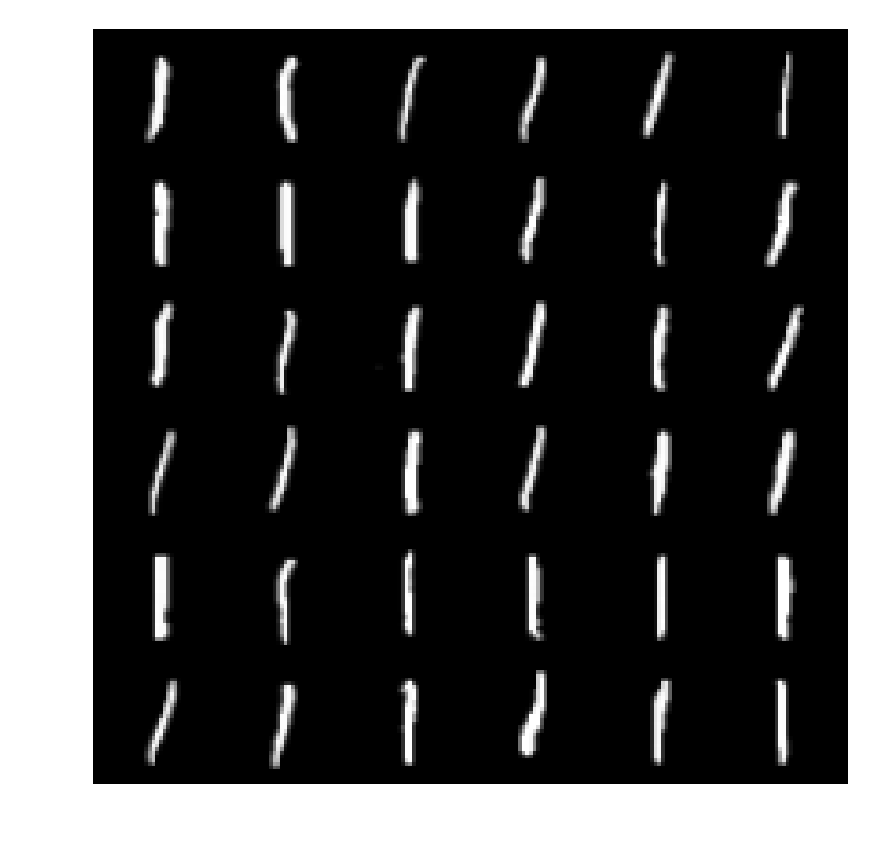} & 
        \includegraphics[width=0.5\columnwidth]{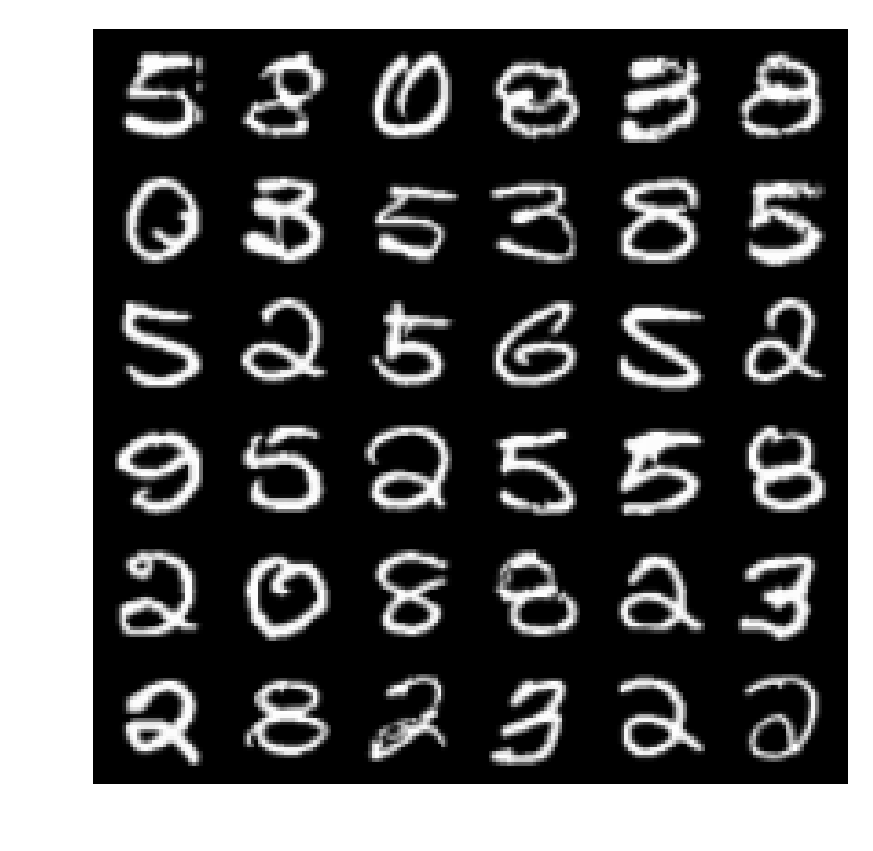} \\
        Most likely MNIST &
        Least likely MNIST
    \end{tabular}
    \caption{Most and least likely MNIST images, as predicted by a regressor for a GAN trained on MNIST without ones}
    \label{fig:mnist_real_top_bottom_no_ones}
\end{figure}

\begin{figure}
\includegraphics[width=\columnwidth]{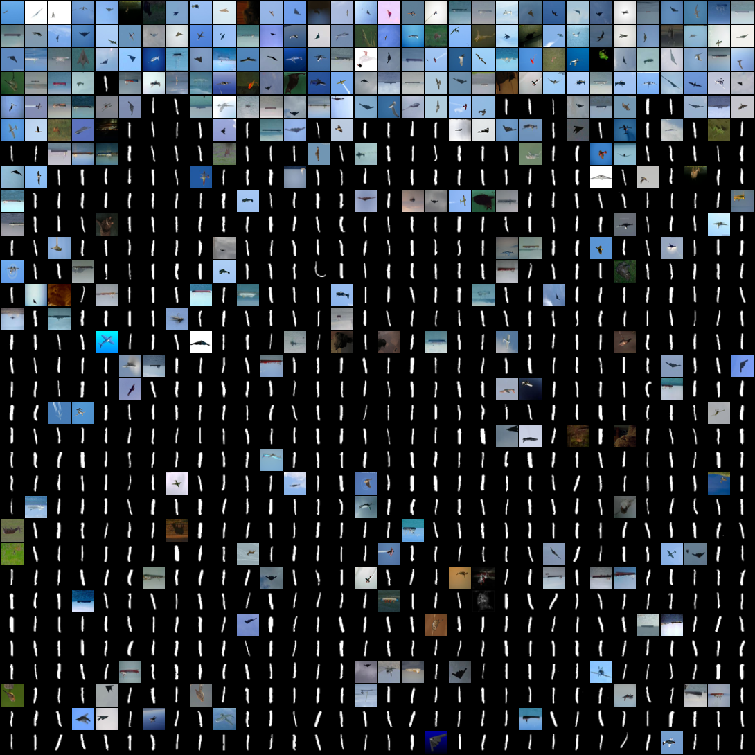}
\caption{Most likely 1024 images from CIFAR and MNIST combined, as predicted by a pixel regressor for a GAN trained only on CIFAR. }
\label{fig:cifar_predicting_mnist}
\end{figure}

\begin{figure}[!htb]
    \centering

    \setlength{\tabcolsep}{2pt} 
    \begin{tabular}{c c}
        \includegraphics[width=0.5\columnwidth]{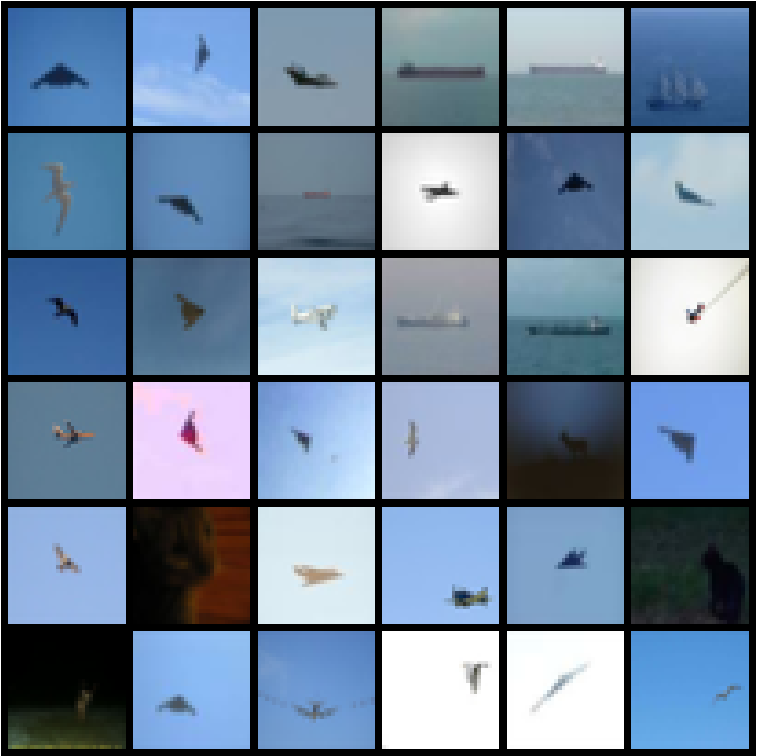} & 
        \includegraphics[width=0.5\columnwidth]{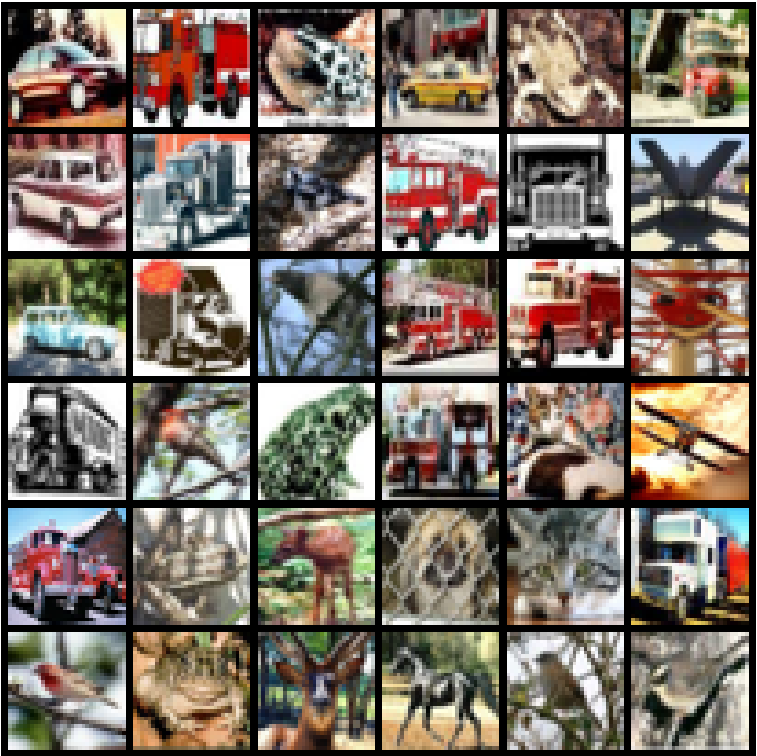} \\
        Top CIFAR &
        Bottom CIFAR
    \end{tabular}
    \caption{Left: Highest probability CIFAR images according to a pixel regressor trained on CIFAR. Right: Lowest probability CIFAR images for the same distribution.}
    \label{fig:cifar_real_top_bottom}
\end{figure}

\begin{figure}[!htb]
    \centering

    \setlength{\tabcolsep}{2pt} 
    \begin{tabular}{c c}
        \includegraphics[width=0.5\columnwidth]{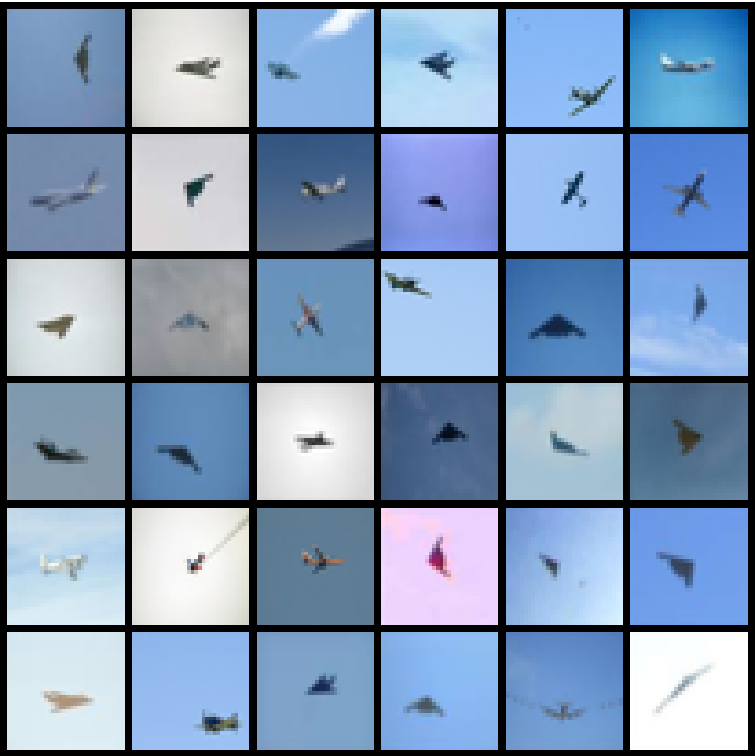} & 
        \includegraphics[width=0.5\columnwidth]{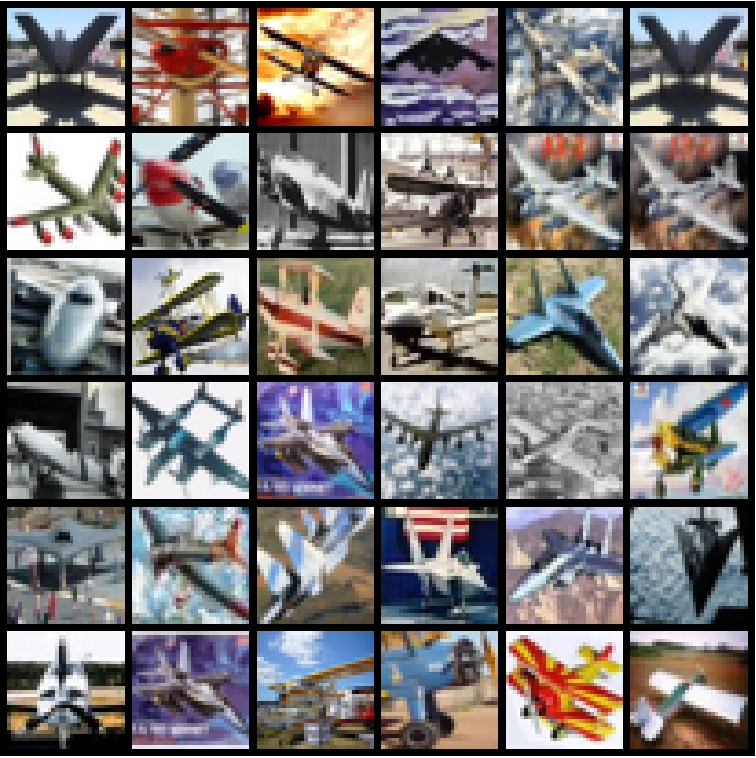} \\
        Top Airplanes &
        Bottom Airplanes
    \end{tabular}
    \caption{Most and least likely CIFAR airplanes, as predicted by a pixel regressor for a CIFAR GAN.}
    \label{fig:cifar_real_airplanes}
\end{figure}

\begin{figure}[!htb]
    \centering

    \setlength{\tabcolsep}{2pt} 
    \begin{tabular}{c c}
        \includegraphics[width=0.5\columnwidth]{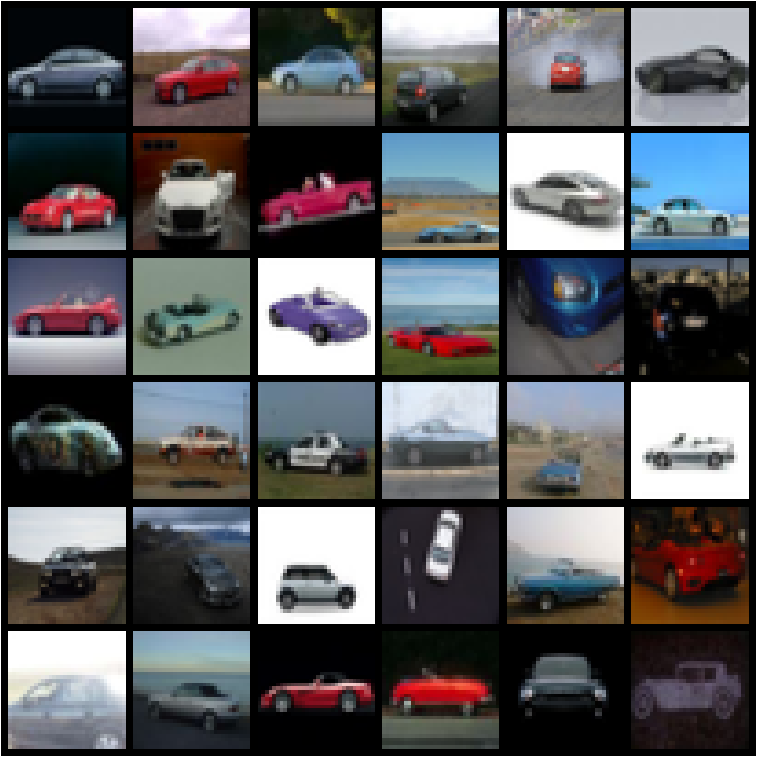} & 
        \includegraphics[width=0.5\columnwidth]{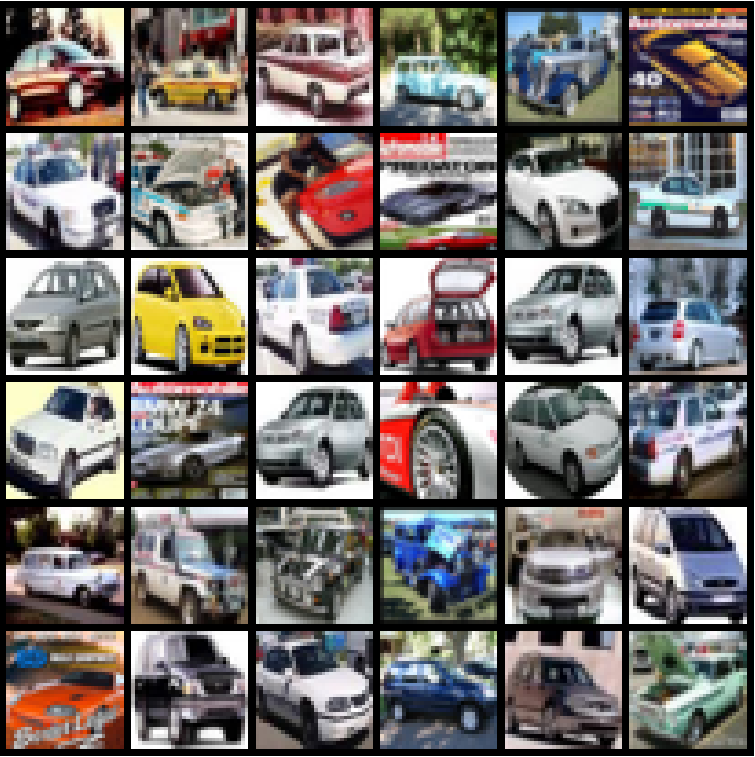} \\
        Top Cars &
        Bottom Cars
    \end{tabular}
    \caption{Most and least likely CIFAR cars}
    \label{fig:cifar_real_cars}
\end{figure}

This section explores the outcomes of image density estimation, and presents difficulties that arise when it is used in practice.  We see that image distribution can be highly irregular and non-uniform -- a characteristic that makes image densities difficult to interpret.

\subsection{Everybody loves 1s}
We saw in Section \ref{compare} that image densities could be used to discern extreme outliers from an image distribution.  But what about the inliers?  In this section we dive further into what image characteristics most strongly determine image density.

Figure \ref{fig:mnist_real_top_bottom} shows the most likely and least likely images from the MNIST dataset.  We see that all of the most likely images are 1s, while all of the least likely images are more ``loopy'' digits.  While this may seem unintuitive at first; 1's are just as likely to occur as any other digits, so why should they have higher probability density? However, this preference for 1s is in fact the result of {\em correct} density estimation;  images of 1s in the MNIST dataset are all very well aligned and similar, and when interpreted as vectors in a high-dimensional space they cluster closely together.  As a result, the 1s define an extremely high-density mode in the image distribution.

To better asses the severity of this problem, we train the density estimator on all images except 1s.  Intuitively, the 1s should now be outliers from the distribution.  However, the density function thinks the opposite.  When the density of this incomplete distribution is evaluated on all MNIST test data (including the 1s), the most likely digits are still 1s (Figure \ref{fig:mnist_real_top_bottom_no_ones}).   

This effect is likely because of the constant black background in images of 1s. Most pixels in these images are black (the most common value), and so these images lie relatively close (in the Euclidean sense) to many other MNIST images,  making them inliers rather than outliers. 

A similar problem is manifested by the CIFAR dataset (Figure \ref{fig:cifar_real_top_bottom}).  In this case, the most likely images contain a simple blue background.  This is likely because the ``airplane'' class contains many images with a smooth background of a similar blue color, and so these images lie close together in the Euclidean distance, defining a high-density mode.  Furthermore, in images of high probability density, the actual object of interest is extremely small and the background is dominant.  Images with large foreground objects contain high-frequency features and do not correlate as well,  so they lie far apart and have relatively low probability density as in Figures \ref{fig:cifar_real_airplanes} and \ref{fig:cifar_real_cars}.


\subsection{Are CIFAR images outliers in their own distribution?} \label{outliers}
Intuitively, one might expect to use density estimates for outlier detection; outliers should have extremely low densities compared to inliers.  We saw in Section \ref{highlow} that densities were able to detect irregular/outlier images produced by a GAN. 

\begin{figure}
\includegraphics[width=\columnwidth]{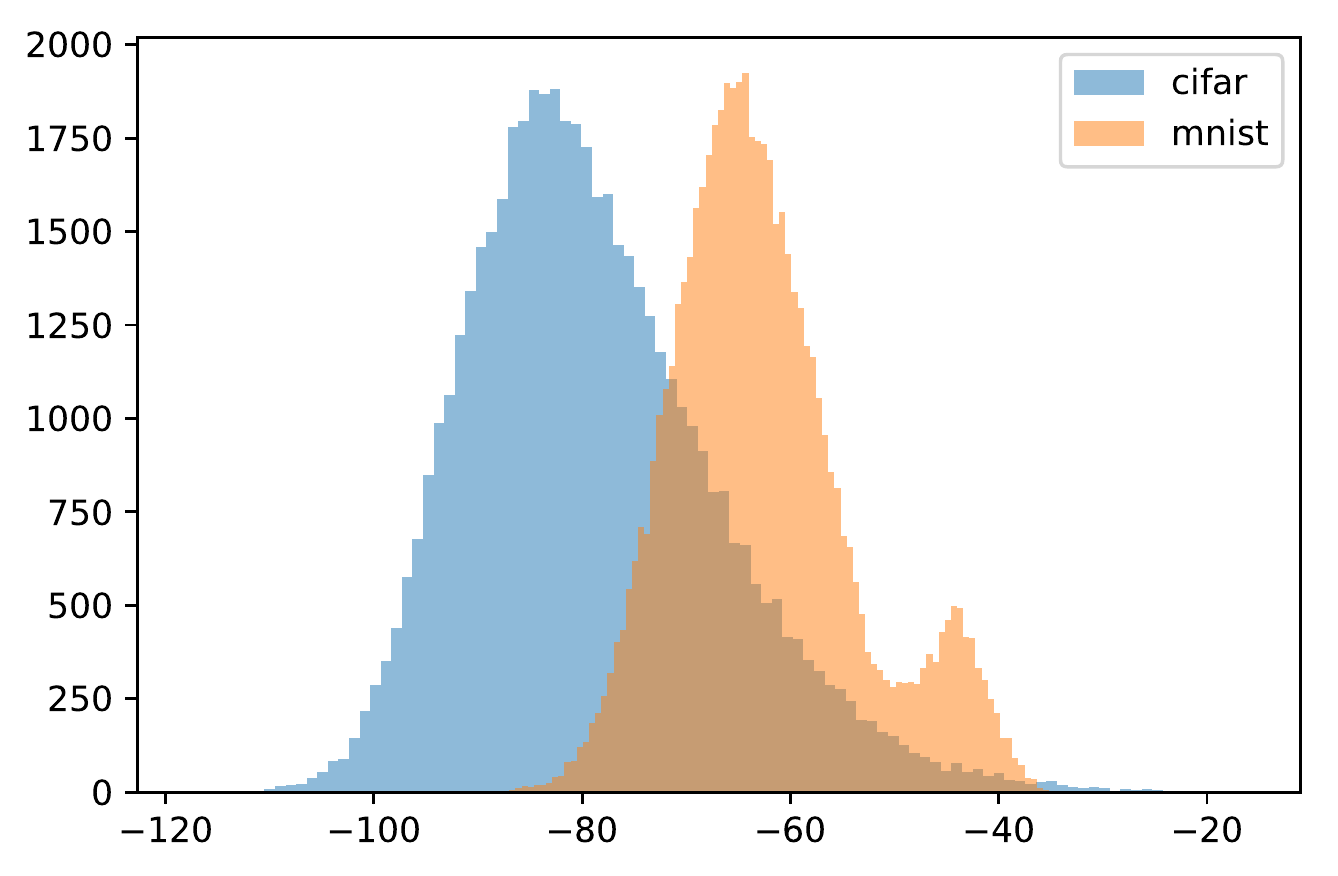}
\caption{Histogram of log probabilities of MNIST and CIFAR, predicted using a pixel-space density estimator for CIFAR.}
\label{mnistwins}
\end{figure}

  We study the seemingly easier task of deciding whether MNIST images are outliers from the CIFAR distribution.   To this end, we train a density model on only CIFAR, and evaluate the density function on both CIFAR and MNIST images.    The most likely images from the combined CIFAR/MNIST dataset are shown in Figure \ref{fig:cifar_predicting_mnist}.   We see that the set of most likely images is dominated by MNIST digits, with a small number of extremely simple CIFAR images in the top as well.   Histograms of these densities are depicted in Figure \ref{mnistwins}, and we see that MNIST is indeed far more likely than CIFAR. 
  
This result is totally consistent with the experiments above -- smooth, geometrically structured images lie close to the center of the distribution.  The MNIST images apparently lie in an extremely high density mode.  However, when we sample the CIFAR distribution, highly structured images of this type seldom appear.  This indicates that the high density region occupies an extremely small volume and thus very small probability mass.  Meanwhile, the lower-density outlying region (which contains the vast majority of the CIFAR images) comprises nearly all the probability mass.

\begin{figure}[!htb]
    \centering

    \setlength{\tabcolsep}{2pt} 
    \begin{tabular}{c c}
        \includegraphics[width=0.45\columnwidth]{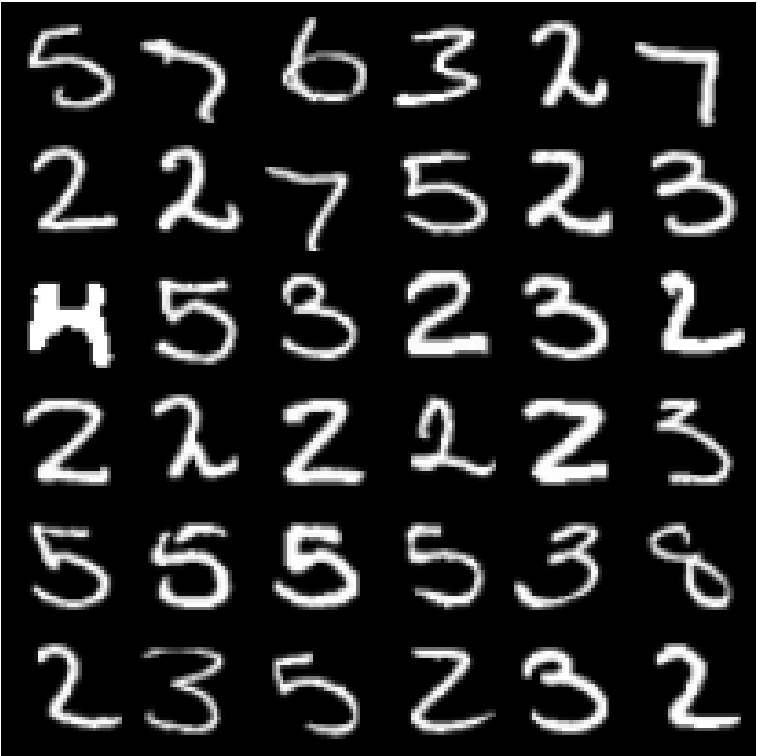} & 
        \includegraphics[width=0.45\columnwidth]{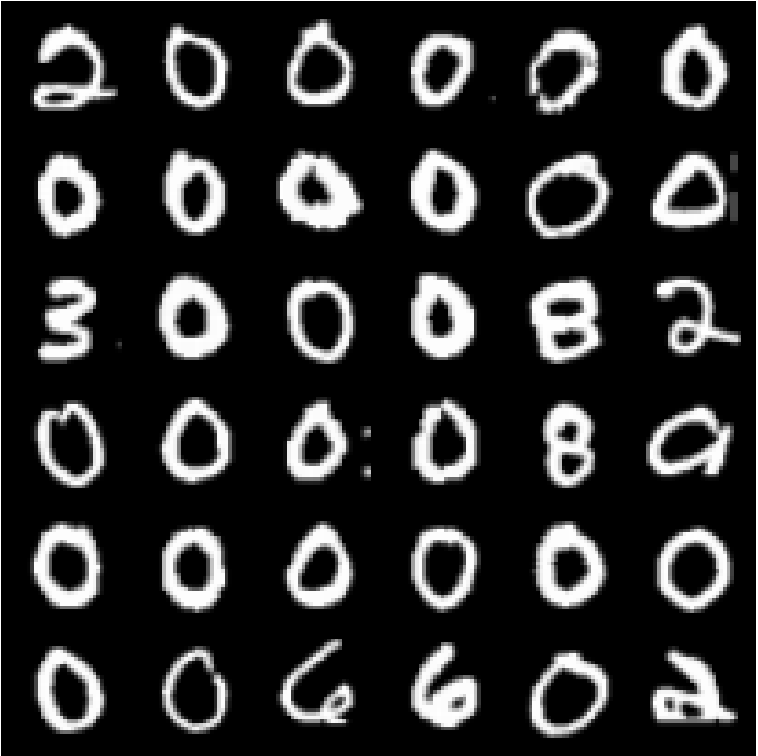} \\
        Top MNIST &
        Bottom MNIST
    \end{tabular}
    \caption{Most and least likely MNIST images, as predicted by the latent code regressor for a GAN trained on MNIST.}
    \label{fig:mnist_real_top_bottom_codes}
\end{figure}



\begin{figure}
\includegraphics[width=\columnwidth]{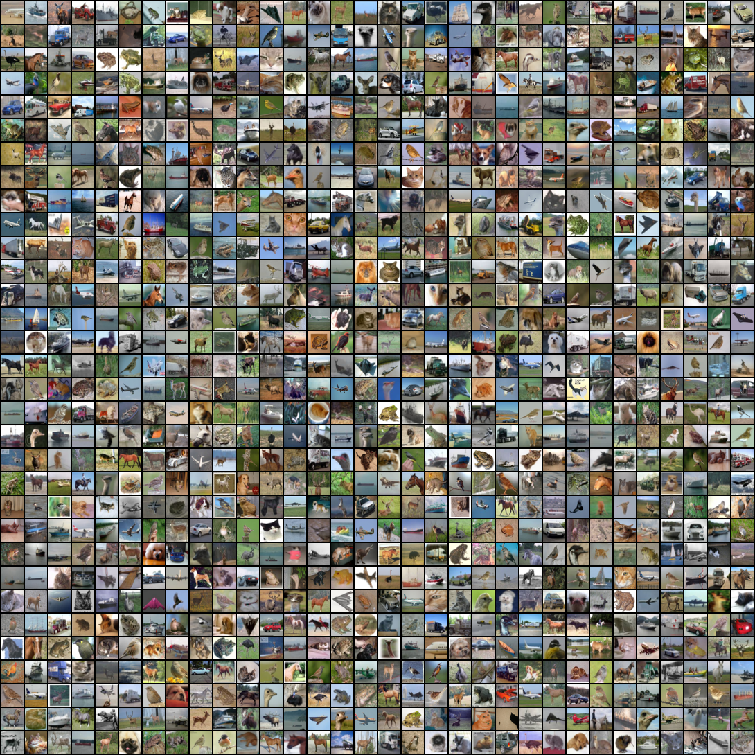}
\caption{Most likely 1024 images from combined CIFAR and MNIST, as predicted by a latent code regressor from a GAN trained on CIFAR.}
\label{fig:cifar_predicting_mnist_codes}
\end{figure}

\begin{figure}[!htb]
    \centering

    \setlength{\tabcolsep}{2pt} 
    \begin{tabular}{c c}
        \includegraphics[width=0.5\columnwidth]{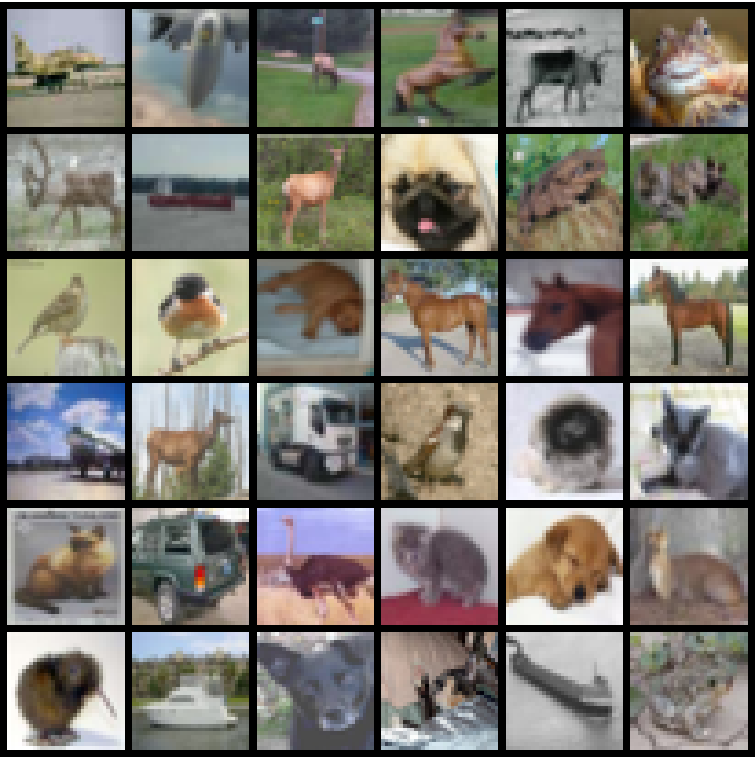} & 
        \includegraphics[width=0.5\columnwidth]{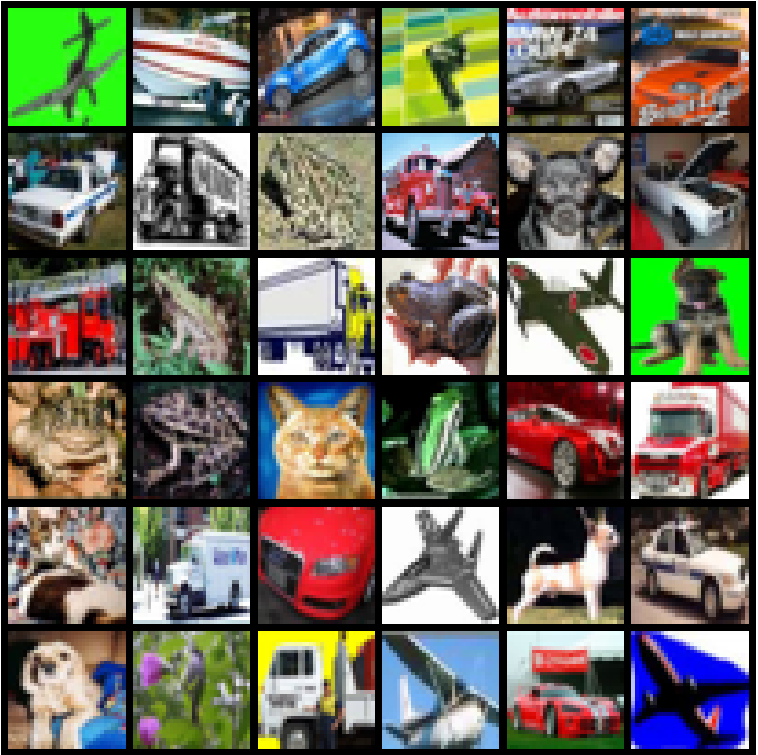} \\
        Top CIFAR &
        Bottom CIFAR
    \end{tabular}
    \caption{Most and least likely CIFAR images, as predicted by a latent code regressor trained on CIFAR.}
    \label{fig:cifar_real_top_bottom_codes}
\end{figure}

\begin{figure}[!htb]
    \centering

    \setlength{\tabcolsep}{2pt} 
    \begin{tabular}{c c}
        \includegraphics[width=0.5\columnwidth]{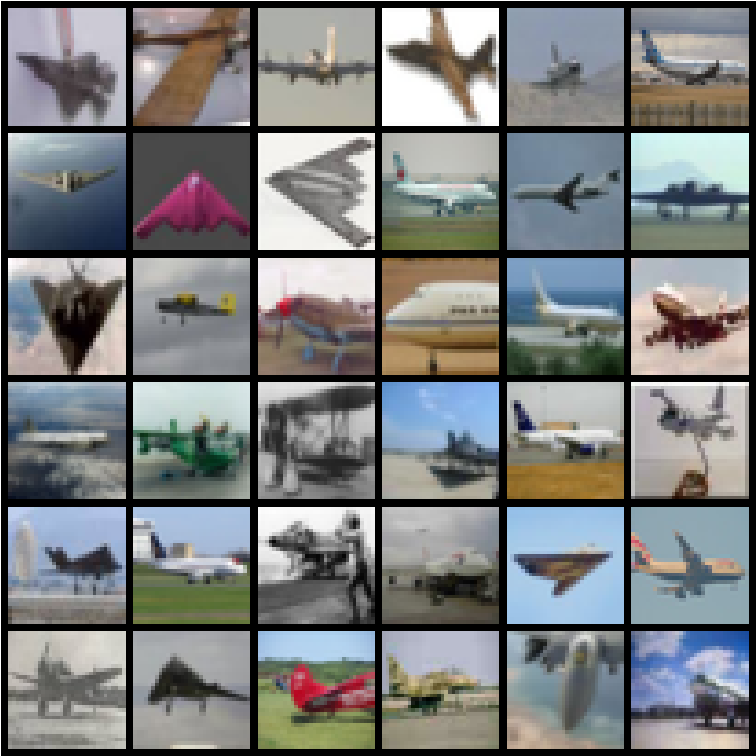} & 
        \includegraphics[width=0.5\columnwidth]{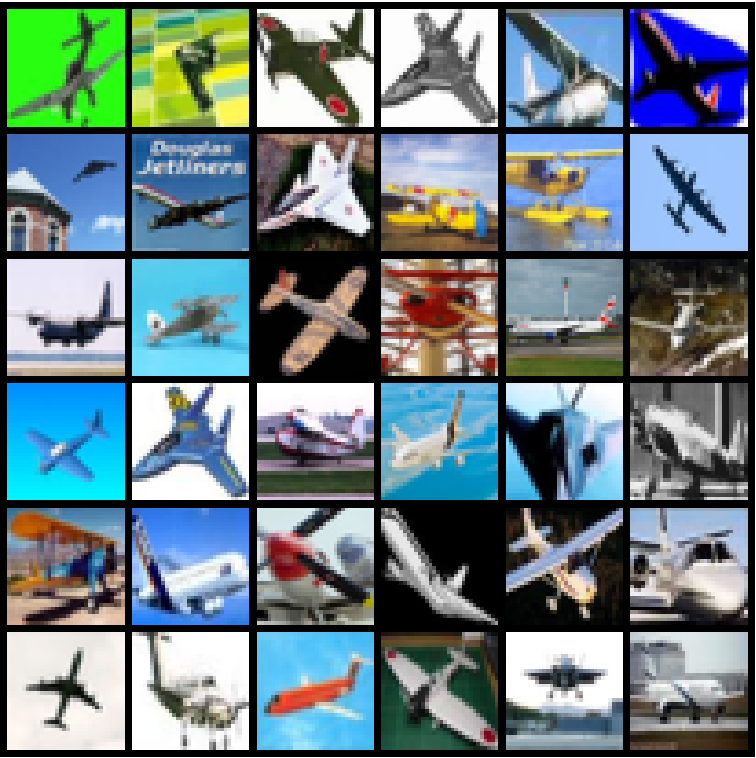} \\
        Top Airplanes &
        Bottom Airplanes
    \end{tabular}
    \caption{Most and least likely CIFAR airplanes, as predicted by a latent code regressor trained on CIFAR.}
    \label{fig:cifar_real_airplanes_codes}
\end{figure}

\section{Making density functions interpretable}

The experiments in Section \ref{wish} indicate that probability densities on complex image datasets have highly unintuitive structure.  Fairly ``typical''  images are often outliers that lie far from the modes of the distribution.   This lack of interpretability is a consequence of a well known problem; the Euclidean distance between images does not capture an intuitive or semantically meaningful concept of similarity.  ``Outliers'' of a distribution are points that lie far from the modes in a {\em Euclidean} sense, and so we should expect the relationship between density and semantic structure to be tenuous.

 To make density estimates interpretable, we need to embed images into a space where Euclidean distance has semantic meaning.  We do this by embedding images into a deep feature space.  In deep feature space, nearby images have similar semantic structure, and well separated images are semantically different.  This enables distributions to have interpretable modes and outliers.

There are many options to choose from when selecting a deep embedding.
In the unsupervised setting where we already have a GAN at our disposal, the simplest choice for a feature embedding is to associate images with their latent representation $z,$ the pre-image of the GAN.  This embedding is particularly simple  because the density function in this space is simply the Gaussian density, which can be evaluated in closed form.  We learn this density mapping by associated each image with the density of its pre-image $z,$ without accounting for the Jacobian of the mapping.



\subsection{Images are now inliers in their own distribution}

\begin{figure}
\includegraphics[width=\columnwidth]{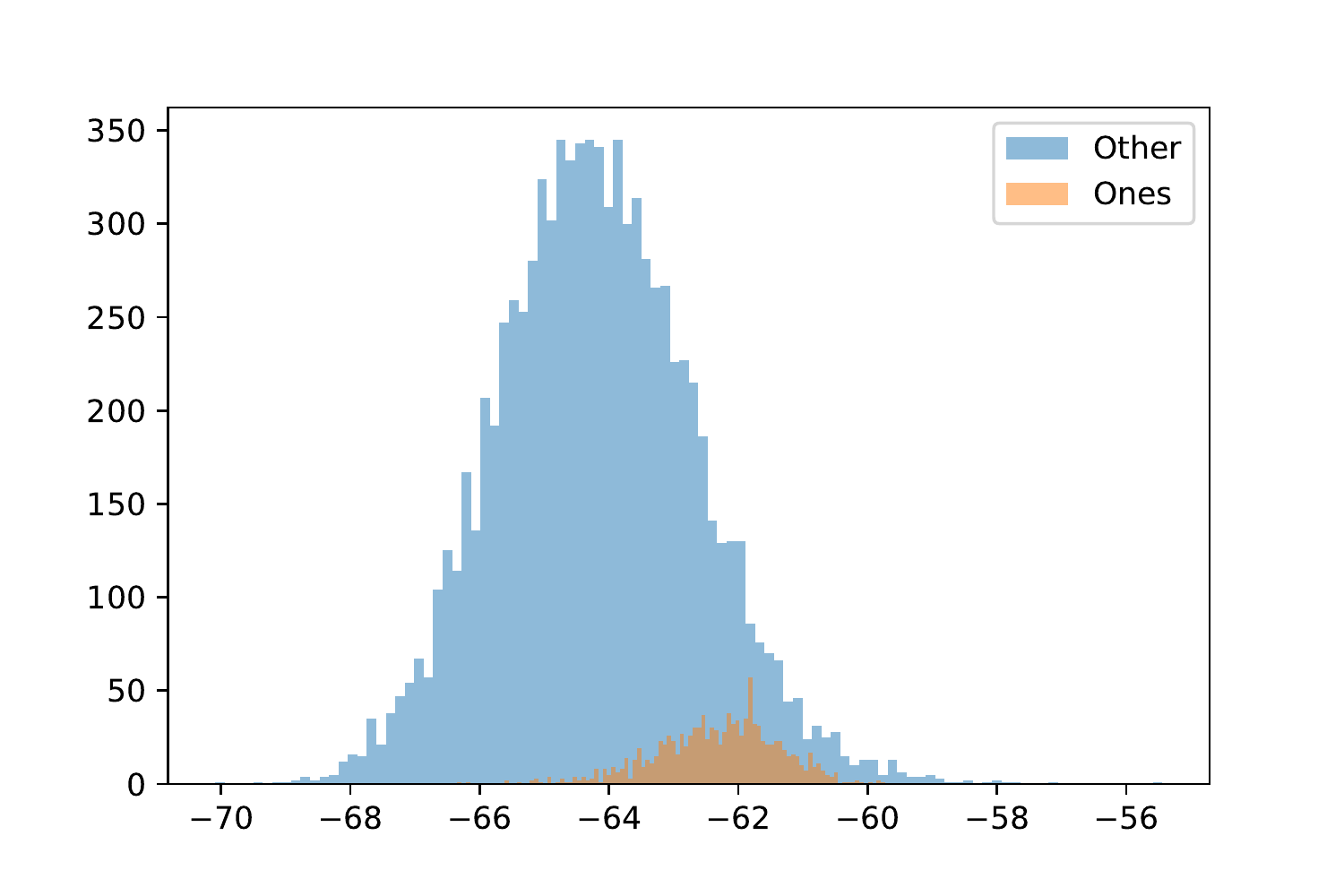}
\caption{Histogram of log probabilities of MNIST, as predicted by a latent code regressor for a GAN trained on MNIST. Note that the log probability values are much more clustered than in pixel space.}
\label{fig:mnist_code_log_probs}
\end{figure}

We show the most and least likely MNIST images under the deep feature model in Figure \ref{fig:mnist_real_top_bottom_codes}.  Unlike the pixel-space model, there is now diversity in the highest digits, and the distribution is not dominated by 1s.  The deep model also produces much more uniform probabilities than the pixel model (Figure \ref{fig:mnist_code_log_probs}) -- this is expected since the MNIST dataset is itself fairly uniform with few semantic outliers.  

\begin{figure}
\includegraphics[width=\columnwidth]{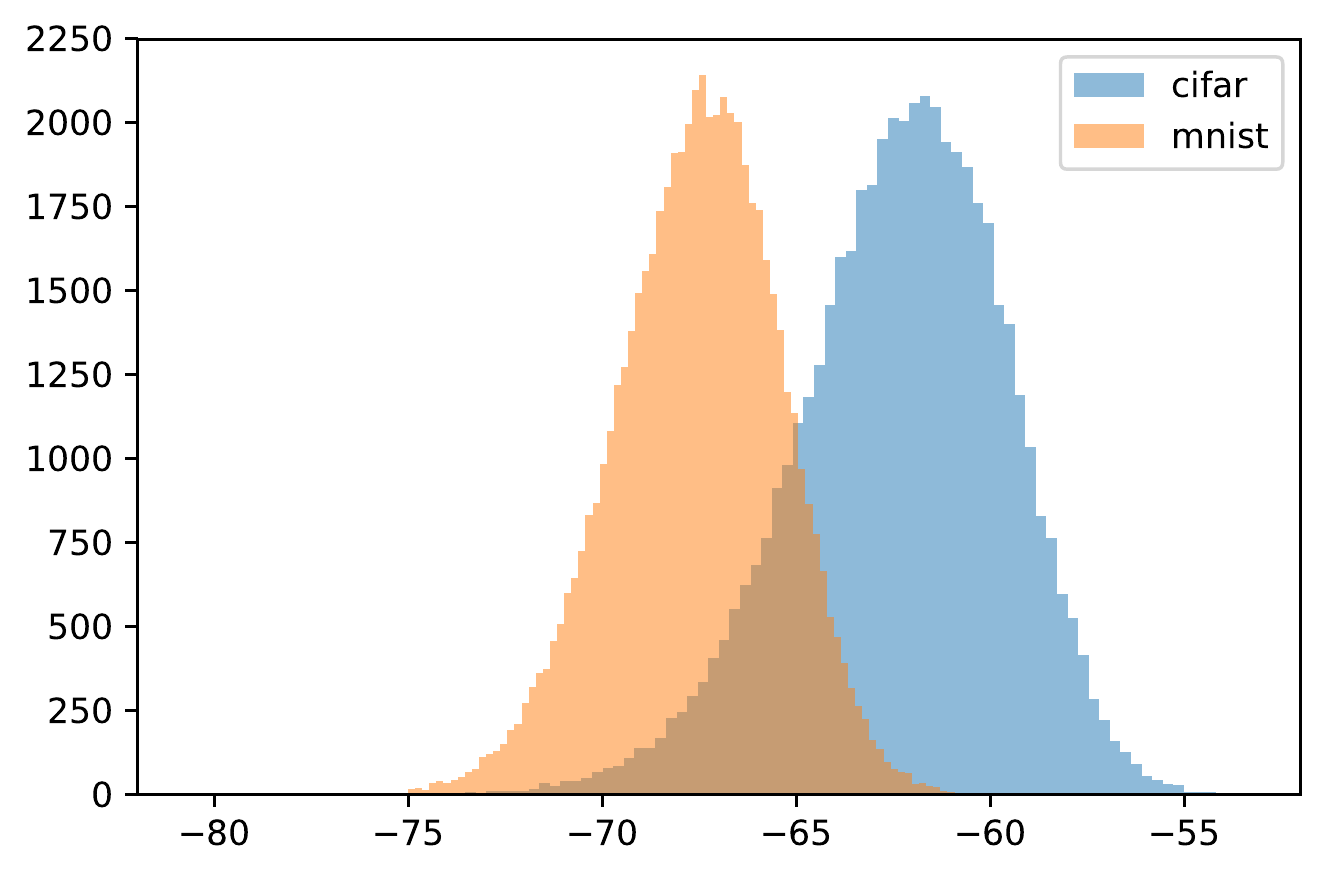}
\caption{Histogram of log probabilities of MNIST and CIFAR, predicted using the latent code regressor for a GAN trained on CIFAR.}
\label{fig:mnist_under_cifar_code_hist}
\end{figure}

We saw in Section \ref{outliers} that MNIST digits were inliers with respect to the CIFAR distribution, and many CIFAR images were outlier in their own distribution (when estimating densities in the pixel space).
When we perform density estimation in deep feature space, density estimates capture a more intuitive notion of outliers.  To show this, we train a deep feature density estimator on the CIFAR distribution only, and then infer probabilities on the combined CIFAR and MNIST dataset.  Figure \ref{fig:mnist_under_cifar_code_hist} shows the histogram of estimated densities.  We see that CIFAR images now occupy high density regions close to the distribution modes, and MNIST images occupy the low density ``outlier'' regions.

Figure \ref{fig:cifar_predicting_mnist_codes} shows the most probable CIFAR and MNIST images with respect to the CIFAR distribution.  Unlike in the case of Figure \ref{fig:cifar_predicting_mnist}, we now see that all of the most likely images are from the CIFAR distribution. 

\subsection{Deep densities depend on image content rather than smoothness}
The most and least likely CIFAR images, according a deep feature density model, are shown in Figure \ref{fig:cifar_real_top_bottom_codes}.  Unlike the pixel-space density estimator depicted in Figure \ref{fig:cifar_real_top_bottom}, the deep feature model favors images where the foreground object is well-defined and occupies a large fraction of the image.  The least likely images contain many objects in unusual configurations (e.g., a car with its hood open), or strange backgrounds (e.g., airplanes with a green sky).

We narrow down to the specific class of airplanes in Figure \ref{fig:cifar_real_airplanes_codes} and we see that the densities no longer depend strongly on the image complexity, but rather on the image content and coloration.   


\section{Conclusion}
Using the power of GANs, we explored the density functions of complex image distributions.  Unfortunately, inliers and outliers of these density functions cannot be readily interpreted as typical and atypical images. However, this lack of interpretability can be mitigated by considering the probability densities not of the images themselves, but of the latent codes that produced them. We postulate that such feature embeddings tend to cluster images in space around more semantically meaningful categories, consolidating probability mass that would otherwise be spread out thinly in pixel space due to many visual variations of the same type of object. There are a host of potential applications for the resulting image PDFs, including detecting outliers and domain shift that will be explored in future work.

\section{Acknowledgements}
We would like to thank the Lifelong Learning Machines program from DARPA/MTO for their support of this project.

{\small
\bibliographystyle{ieee}
\bibliography{bibliography}
}

\end{document}